%% file: main.tex
\documentclass[10pt,twocolumn,letterpaper]{article}

\usepackage{cvpr}              
\usepackage{multirow}
\usepackage{xspace}
\usepackage{caption}
\usepackage{subcaption}
\input{preamble}

\definecolor{cvprblue}{rgb}{0.21,0.49,0.74}
\usepackage[pagebackref,breaklinks,colorlinks,allcolors=cvprblue]{hyperref}
\def\paperID{372} 
\def\confName{CVPR}
\def\confYear{2025}

\usepackage{xcolor}
\usepackage[normalem]{ulem} 

\usepackage[ruled,vlined]{algorithm2e}
\usepackage{algorithmic}

\title{\name: Capturing Temporal Relationships \\ of Deformable 3D Gaussians for Robust Reconstruction

}

\author{Dadong Jiang$^{1,3}$
\and
Zhihui Ke$^1$
\and
Xiaobo Zhou$^{1 \dagger}$
\and
Zhi Hou$^{2 \dagger}$
\and
Xianghui Yang$^{3}$
\and
Wenbo Hu$^{4}$
\and
Qiu Tie$^{1}$
\and
Chunchao Guo$^{3}$
\and
$^1$ Tianjin University 
$^2$ Shanghai Artificial Intelligence Laboratory \\
$^3$ Tencent Hunyuan
$^4$ Tencent AI Lab \\
}

\begin{document}

\makeatletter
\g@addto@macro\@maketitle{
\vspace{-8mm}
  \includegraphics[width=1.0\textwidth, trim=0 0 0 0, clip]
  {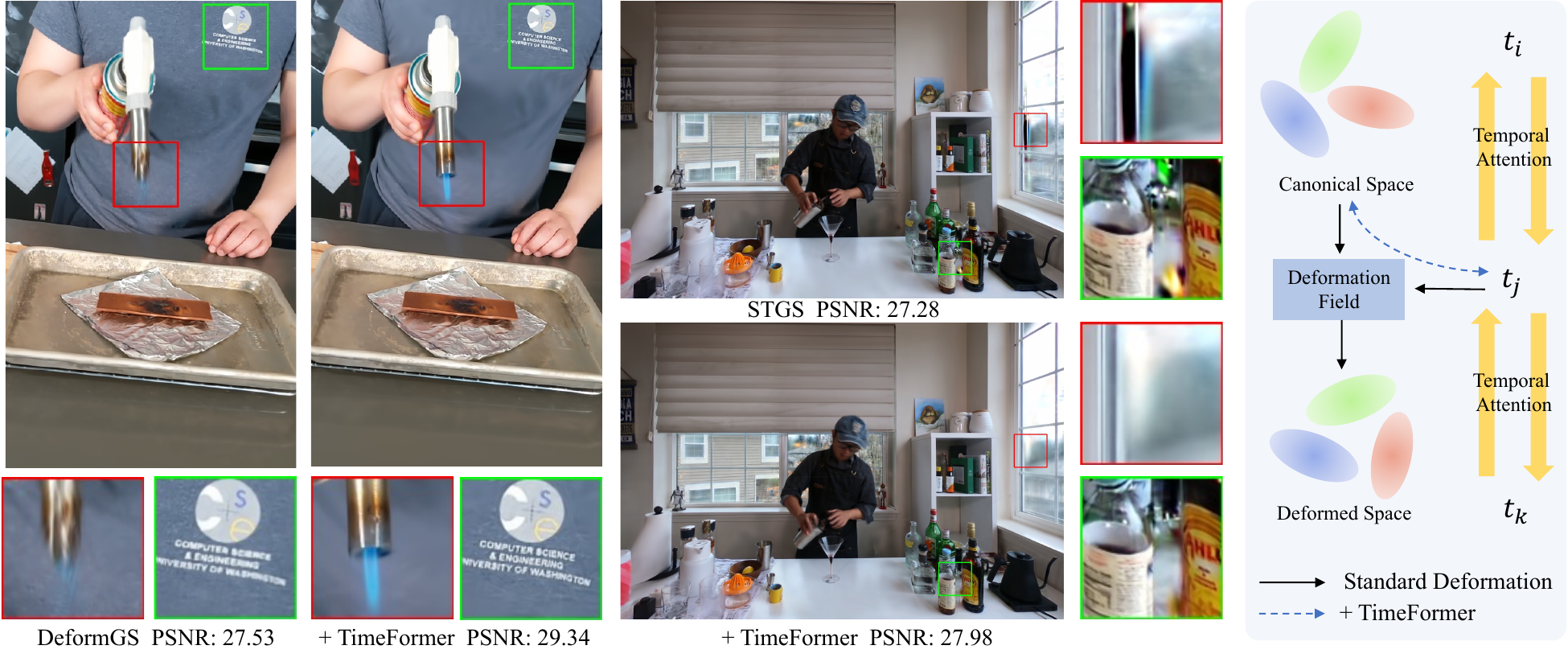}
  \vspace{-8mm}
  \captionof{figure}{We propose \textbf{\name}, a Transformer module that implicitly models the motion pattern via Temporal Attention from a learning perspective (right). \name is plug-and-play to existing deformable 3D Gaussian reconstruction methods~\cite{Wu_2024_CVPR, Li_STG_2024_CVPR, yang2023deformable3dgs} and enhances reconstruction results (left and middle) without affecting their original inference speed.}
  \label{fig: teaser}
  \vspace{5mm}
}
\makeatother
\maketitle

\let\thefootnote\relax\footnotetext{$\dagger$ Corresponding Author}
\input{sec/0_abstract}  

\input{sec/1_intro}

\input{sec/2_related_works}
\input{sec/3_preliminary}

\input{sec/4_method}
\input{sec/5_experiment}

\input{sec/6_conclusion}
{
    \small
    \bibliographystyle{ieeenat_fullname}  
    \bibliography{main}
}

\input{sec/X_suppl}

\end{document}

%% file: preamble.tex
\newcommand{\name}{TimeFormer\xspace}
\newcommand{\encoder}{Cross-Temporal Encoder\xspace}

\definecolor{commentcolor}{RGB}{110,154,155}   
\newcommand{\PyComment}[1]{\ttfamily\textcolor{commentcolor}{\# #1}}  
\newcommand{\PyCode}[1]{\ttfamily\textcolor{black}{#1}} 

%% file: sec/0_abstract.tex
\begin{abstract}

Dynamic scene reconstruction is a long-term challenge in 3D vision. Recent methods extend 3D Gaussian Splatting to dynamic scenes via additional deformation fields and apply explicit constraints like motion flow to guide the deformation. However, they learn motion changes from individual timestamps independently, making it challenging to reconstruct complex scenes, particularly when dealing with violent movement, extreme-shaped geometries, or reflective surfaces.
To address the above issue, we design a plug-and-play module called \name to enable existing deformable 3D Gaussians reconstruction methods with the ability to implicitly model motion patterns from a learning perspective.
Specifically, \name includes a Cross-Temporal Transformer Encoder, which adaptively learns the temporal relationships of deformable 3D Gaussians.
Furthermore, we propose a two-stream optimization strategy that transfers the motion knowledge learned from \name to the base stream during the training phase. This allows us to remove \name during inference, thereby preserving the original rendering speed.
Extensive experiments in the multi-view and monocular dynamic scenes validate qualitative and quantitative improvement brought by \name.
Project Page: \href{https://patrickddj.github.io/TimeFormer/}{https://patrickddj.github.io/TimeFormer/}
\end{abstract}

%% file: sec/1_intro.tex
\section{Introduction}
\label{sec:intro}

High-quality reconstruction of dynamic scenes is significantly challenging in computer vision and graphics, yet has a wide range of potential applications in movie production, virtual reality, and augmented reality. The difficulty stems from factors like occlusions, translucent materials, specular surfaces, and changing topology, all of which are prevalent in dynamic scenes.


Inspired by the success of neural radiance field~(NeRF) on static scenes~\cite{mildenhall2021nerf,barron2021mip,muller2022instant}, extending NeRF to dynamic scenes~\cite{li2022streaming,tian2023mononerf,zhou2024dynpoint,park2021hypernerf,park2021nerfies,liu2023robust,li2021neural,gao2021dynamic} has been explored. However, these NeRF-based methods are limited by computationally intensive volume rendering~\cite{drebin1988volume}, which makes real-time rendering almost impossible.
Recently, 3D Gaussian Splatting~(3DGS)~\cite{kerbl3Dgaussians} represents the scene with anisotropic 3D Gaussians and develops a rasterization-based rendering algorithm,  which allows real-time rendering through directly projecting 3D Gaussians onto the image plane. Following the revolutionized 3DGS, dynamic Gaussian Splatting methods~\cite{Li_STG_2024_CVPR,4drotorgs,yang2023gs4d,sun20243dgstream} have been introduced for reconstructing dynamic scenes. These methods typically construct a canonical 3DGS and utilize a deformation field to deform it based on individual timestamps~\cite{yang2023deformable3dgs, Li_STG_2024_CVPR, Wu_2024_CVPR}.

\begin{figure}[tbp]
    \centering
    \includegraphics[width=\linewidth, trim=20 0 20 20, clip]{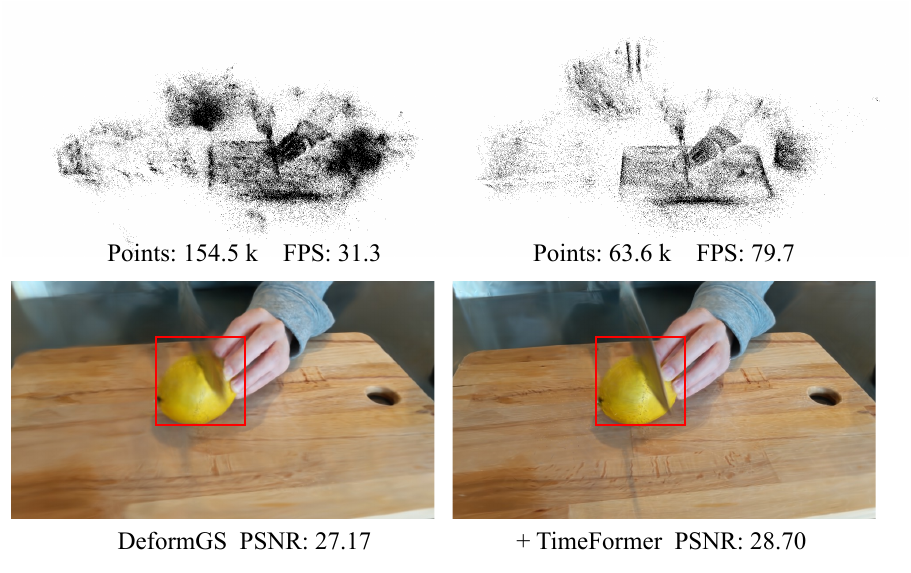}
    \vspace{-8mm}
    \caption{\name guides towards more efficiently distributed canonical space, showing higher FPS and better quality.  The results are from ``cut lemon'' in the HyperNeRF Dataset~\cite{park2021hypernerf}.}
    \label{fig: point_cloud}
    \vspace{-5mm}
\end{figure}

However, temporal relationships of 3D Gaussians have been poorly investigated. Previous studies have primarily modeled motion patterns using various types of the deformation field, for example, MLPs~\cite{luiten2024dynamic, bae2024per, lu20243d, shaw2024swings, yang2023deformable3dgs, huang2024sc,zhao2024gaussianprediction,wansuperpoint}, spatial-temporal planes~\cite{Wu_2024_CVPR, lu2024dn, duisterhof2023md}, polynomial functions~\cite{Li_STG_2024_CVPR}, Fourier series~\cite{katsumata2024a}, and combinations of these methods~\cite{lin2024gaussian}.
These methods learn motion patterns from independent time input in a vanilla way, neglecting internal cross-time relationships.
Some studies introduce the motion flow regularization~\cite{Li2024ST} to explicitly learn motion patterns from neighboring frames~\cite{luiten2024dynamic,lu2024dn}. 
Although these methods promote similar motion between adjacent timestamps, they adopt a local perspective on time series during optimization.
This constraint makes it challenging to reconstruct scenes with more complex motion patterns, such as sudden appearances, violent movements, or reflective surfaces. 
In this paper, we introduce \name, a transformer module designed to implicitly learn motion patterns across multiple timestamps.
By modeling temporal relationships within a time batch, we aim to provide a global view of the entire time series, enabling the deformation field backbones themselves to capture motion patterns from a learning perspective, as shown in Fig.~\ref{fig: teaser}.
Specifically, \name utilizes a \encoder to capture the implicit motion patterns of 3D Gaussians across multiple sampled timestamps using a self-attention mechanism.
Moreover, to avoid additional computational costs of \name during inference, we present a two-stream optimization strategy. By sharing the weights of two deformation fields, we transfer the motion pattern learning from \name stream to the base stream during training. Therefore, we can eliminate \name during inference and maintain the original rendering speed. 
Notably, \name does not require any prior information and extracts motion patterns solely from RGB supervision, which can be seamlessly adapted to previous deformable 3D Gaussian methods in a plug-and-play manner.
Furthermore, \name promotes faster gradient descent and guides a more efficient canonical space, ultimately increasing FPS during inference, as demonstrated in Fig.~\ref{fig: point_cloud}.





To sum up, our main contributions are as follows:
\begin{itemize}
    \item We propose \name, a transformer module that enhances current deformable 3D Gaussians reconstruction methods in a plug-and-play manner from an automatic learning perspective.
    \item The two-stream optimization strategy allows the exclusion of \name during inference while maintaining and even improving rendering speed.
    \item Extensive experiments on real-world datasets validate the effectiveness of \name, achieving state-of-the-art rendering quality.
\end{itemize}


%% file: sec/2_related_works.tex
\section{Related Works}
\label{sec:formatting}

\subsection{Dynamic Scene Reconstruction}
Dynamic Scene Reconstruction has been extensively researched over many years, with a wealth of studies~\cite{kanade1997virtualized,zitnick2004high,park2021hypernerf,yang2023deformable3dgs,yang2002real} contributing to current progress. The pioneering neural radiance field~\cite{mildenhall2021nerf,barron2021mip,yu2021plenoctrees,sun2022direct,fridovich2022plenoxels} has demonstrated photorealistic rendering for novel view synthesis from calibrated multiview images, inspiring extensive approaches~\cite{du2021neural,xian2021space,tretschk2021non, pumarola2021d,park2021nerfies, fang2022fast, park2021hypernerf} to extend NeRF to 4D space-time field for dynamic scene reconstruction. There are majorly two lines of NeRF-based methods for dynamic scene reconstruction: 1) deformation-based methods~\cite{tretschk2021non, pumarola2021d,park2021nerfies, fang2022fast, park2021hypernerf} model deformation changes by using a deformation field to map the queried positions in different timestamps to a canonical space; 2) compact 4D space-time fields~\cite{du2021neural,xian2021space,li2022nv3d,chen2022tensorf} take the timestamps, positions, and directions as input to predict color and density. However, NeRF-based approaches are limited to training and rendering speed. Thus, a large number of methods~\cite{li2022streaming, liu2022devrf, wang2022mixed, wang2023neural, shao2023tensor4d, fridovich2023k, cao2023hexplane, xu20244k4d, wu2024tetrirf, wang2024masked, song2023nerfplayer, attal2023hyperreel, lou2024darenerf} have been proposed to accelerate NeRF-based methods. 

Recent 3DGS~\cite{kerbl3Dgaussians} achieves real-time rendering and dynamic 3DGS approaches~\cite{yang2023gs4d, yang2023deformable3dgs, Li_STG_2024_CVPR} emerge. In details, deformation-based methods~\cite{luiten2024dynamic, bae2024per, lu20243d, shaw2024swings, yang2023deformable3dgs, huang2024sc, zhao2024gaussianprediction, wansuperpoint, Wu_2024_CVPR, lu2024dn, duisterhof2023md, Li_STG_2024_CVPR, katsumata2024a, lin2024gaussian} employ a deformation field to predict per-Gaussian offsets and deform a canonical 3DGS according to different timestamps. Meanwhile, 4DGS~\cite{yang2023gs4d} and 4D-Rotor~\cite{4drotorgs} add the time dimension to 3D Gaussian, forming a 4D Gaussian representation. TimeFormer focuses on modeling the temporal relationship and is plug-and-play to deformation-based methods~\cite{yang2023deformable3dgs} and 4D space-time representations~\cite{yang2023gs4d,Li_STG_2024_CVPR}. There are also a few attempts~\cite{sun20243dgstream,shaw2024swings,lin2024gaussian} in leveraging the nearby motion or flow for dynamic Gaussian reconstruction. For example, 3DGStream~\cite{sun20243dgstream} proposes a per-frame optimization strategy, which predicts current Gaussian attributes based on previous Gaussian attributes. Meanwhile, Gaussian-Flow~\cite{lin2024gaussian} presents explicitly model time-dependent residual of each attribute. To model a long video sequence, SWinGS~\cite{shaw2024swings} splits a video sequence into multiple sliding windows based on optical flow and uses a deformation-based method to model each sliding window. 

\subsection{Motion Modeling}

Motion prediction in dynamic scene reconstruction from multi-view videos or monocular videos is an ill-posed problem. Early dynamic NeRF~\cite{gao2021dynamic, tretschk2021non, park2021nerfies, park2021hypernerf} directly learns motion patterns from individual timestamps 
, and several later works propose motion flow regularization terms~\cite{gao2021dynamic, li2021neural, song2022pref, tian2023mononerf, liu2023robust, park2023temporal, wang2023flow, somraj2024factorized, zhan2024kfd, miao2024ctnerf} to promote the learning of cross-time motion patterns. These methods typically utilize 2D prior information (\eg, optical flow) from pre-trained networks~\cite{teed2020raft} to supervise the scene flows within neighboring frames. PREF~\cite{song2022pref} uses motion predictor to infer current motion based on the previous four frames and propose a self-supervision strategy without optical flow prior. KFD-NeRF~\cite{zhan2024kfd} models motion patterns by Kalman Filter based on the previous two frames. CT-NeRF~\cite{miao2024ctnerf} employs a cross-attention mechanism from transformer~\cite{vaswani2017attention} to learn the correlation of conservative frames.

Dynamic 3D Gaussian Splatting (3DGS) methods~\cite{guo2024motion, wang2024gflow, gao2024Gaussianflow, zhu2024motiongs} also utilize optical flow supervision to enhance reconstruction performance. Techniques such as MD-Splatting~\cite{duisterhof2023md}, D3DG~\cite{luiten2024dynamic}, and ST-4DGS~\cite{Li2024ST} focus on minimizing both forward and backward Gaussian flow to promote temporal smoothness in Gaussian motions. DN-4DGS~\cite{lu2024dn} incorporates information from previous, current, and next timestamps into the deformation field to capture cross-time motion patterns. However, these methods primarily establish temporal correlations with neighboring timestamps
, which limits their ability to address long-term motion changes. Additionally, they introduce extra computational costs during inference, which can reduce rendering speed. In contrast, \name captures the temporal relationships of Gaussians from a global perspective across the entire time series.
Notably, \name is employed only during the training phase, without any additional computational costs during inference.


%% file: sec/3_preliminary.tex
\section{Preliminary: Deformable 3D Gaussians}
3D Gaussians~\cite{kerbl3Dgaussians} represent the scene with a set of 3D Gaussians, each of which has unique opacity $o \in [0,1]$, center position $\mu \in \mathbb{R}^{3\times1}$, and covariance matrix $\Sigma \in \mathbb{R}^{3\times3}$. 
For a position $x \in \mathbb{R}^{3\times1}$ in 3D space, the corresponding contribution of a 3D Gaussian on it can be formulated as:
\begin{equation}
\label{equ: contri_gs}
    G(x) = o \cdot e^{ -\frac{1}{2} (x-\mu)^{\top} \Sigma^{-1} (x-\mu) }.
\end{equation}
The covariance matrix $\Sigma$ can be decomposed into a scaling matrix $\mathbf{S}$ and a rotation matrix $\mathbf{R}$:
    $\Sigma = \mathbf{R}\mathbf{S}\mathbf{S}^T\mathbf{R}^T$,
where $\mathbf{S} = \text{diag}([s_x, s_y, s_z])$ and $\mathbf{R}$ can be transformed from a quaternion $[r_w, r_x, r_y, r_z]$.
Then the 3D Gaussians can be splatted to a 2D camera plane through differential Gaussian splatting.
To model the appearance of 3D Gaussians, spherical harmonics (SH) are introduced to define the color $c$.
Finally, for each pixel, the rendering results of 3DGS can be derived by calculating the color contribution of all the related Gaussians. This process is known as $\alpha$-blending:
\begin{equation}
\label{formula: splatting&volume rendering}
    C = \sum_{i}^N c_i \alpha_i \prod_{j=1}^{i-1} (1-\alpha_j),
\end{equation}
where $c_i$, $\alpha_i$ represent the color and density computed from the $i$-th 3D Gaussian.

Recent methods~\cite{yang2023deformable3dgs, Wu_2024_CVPR, Li_STG_2024_CVPR} utilize deformation fields that extend 3DGS to 4D space, inspired from NeRF-based methods such as D-NeRF~\cite{pumarola2021d}. The structure of the deformation field $\mathcal{D}$ can vary among MLP~\cite{yang2023deformable3dgs}, K-Plane~\cite{Wu_2024_CVPR} and Polynominal~\cite{Li_STG_2024_CVPR}, while the deformation process can be summarized as follows:
\begin{equation}
    (\Delta \mu, \Delta r, \Delta s) = \mathcal{D}(\mu, t) , 
    \label{eq: deform}
\end{equation}
where timestamp $t \in \mathcal{T}$, $\mathcal{T} \in \mathbb{R}^{T \times 1}$ contains $T$ linear time inputs, $\mu, r, s$ are the center position, rotation quaternion, and scaling factors of 3D Gaussians, and $\Delta \mu, \Delta r, \Delta s$ are their residuals, respectively. The inputs and outputs of Eq.~\ref{eq: deform} can vary among different methods, while they share the same framework.

%% file: sec/4_method.tex
\begin{figure*}[tbp]
    \centering
    \includegraphics[width=\linewidth]{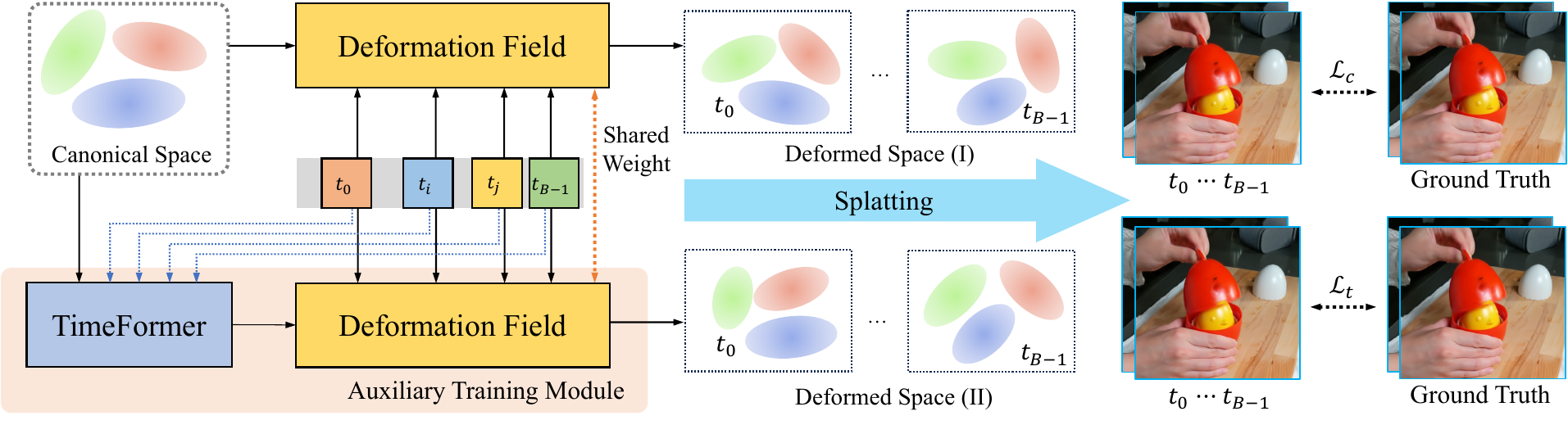}
    \vspace{-8mm}
    \caption{The Framework of Deformable 3D Gaussians Reconstruction with \name. Existing deformable 3D Gaussians framework usually includes the canonical space and the deformation field (first row), we incorporate \name to capture cross-time relationships and explore motion patterns implicitly (second row). 
    We share weights of two deformation fields to transfer the learned motion knowledge. This allows us to \textbf{exclude this Auxiliary Training Module}
 during inference.
    }
    \label{fig: framework}
    \vspace{-5mm}
\end{figure*}

\begin{figure}[tbp]
    \centering
    \includegraphics[width=0.75\linewidth]{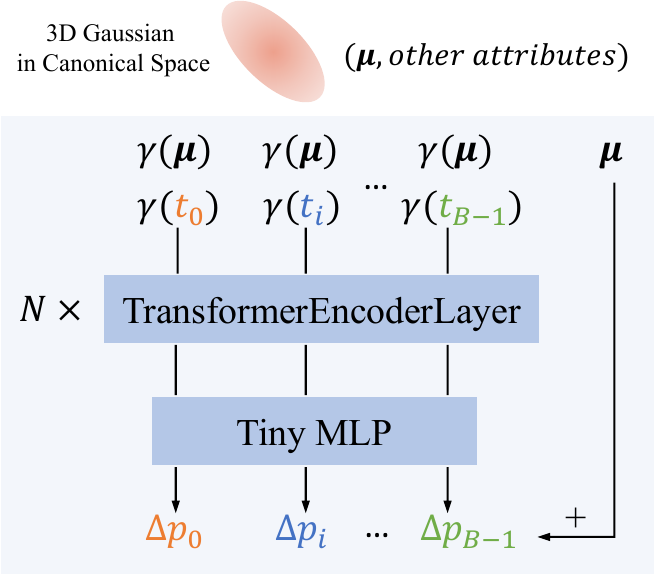}
    \caption{The Structure of \encoder. We concatenate randomly sampled timestamps to position $\mu$, treating them as special \textit{tokens} in a sequence. 
    This module is designed to model multi-temporal relationships and produce distinct time-variant position offsets $\Delta p_0, \dots, \Delta p_{B-1}$.
    }
    \label{fig: gs_timeformer}
    \vspace{-5mm}
\end{figure}

\begin{figure}[tbp]
    \centering
    \includegraphics[width=0.75\linewidth]{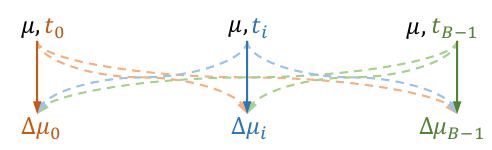}
    \vspace{-3mm}
    \caption{Data Flow Changes in the Deformation Field. Dashed lines represent new data flow among time samples $t_0, \dots,t_{B-1}$.}
    \label{fig: flow}
    \vspace{-5mm}
\end{figure}

\section{Method}
In this section, we first provide an overview of the enhanced deformable 3D Gaussians reconstruction with the proposed \name (Sec.~\ref{sec: overview}). We then introduce the \name in detail (Sec.~\ref{sec: gs_timeformer}) which consists of a \encoder and shared deformation fields (Sec.~\ref{sec: shared}). We also discuss the insights behind \name from the view of gradient flow.

\subsection{Overview}
\label{sec: overview}
Previous methods model motion patterns by explicitly learning temporal relationships on individual or neighboring timestamps, failing on those complex scenes containing violent movement or dynamic reflective surfaces. In contrast, we present \name to enable the deformable 3D Gaussian backbones themselves to model cross-time relationships from an implicit learning perspective. The main framework with the proposed \name is shown in Fig.~\ref{fig: framework}. 
Our approach retains standard reconstruction modules, which include (1) 3D Gaussians in the canonical space and (2) a deformation field that applies time-variant transformation. Additionally, \name is introduced before the deformation field to extract implicit cross-time motion features for each Gaussian through a self-attention mechanism along the time dimension. 
Moreover, we share the weights of two deformation fields to transfer the motion knowledge from \name to mitigate the gap between the original branch and \name branch, which supports real-time rendering without \name during inference.

\subsection{\name}
\label{sec: gs_timeformer}

\noindent \textbf{\encoder}
BatchFormer~\cite{hou2022batch} demonstrates that the attention mechanism helps learn sample relationships from batch dimension, rather than channel and spatial dimentions~\cite{vaswani2017attention, dosovitskiy2020vit}, inspiring us that different timestamps can also be considered as a special time batch. Let $\mathcal{T}_s \subset \mathcal{T}, \mathcal{T}_s = \{t_i\}_{i=0}^{B-1}$ denotes randomly sampled timestamps, and let $\mathcal{G} \in \mathbb{R}^{N \times (3+C)}$ denotes Gaussians in the canonical space, where $B$ is the size of time batch, $N$ is the number of Gaussians, $3+C$ means each Gaussian has 3 position channels and $C$ additional channels. 
As in Fig.~\ref{fig: gs_timeformer}, all Gaussians's postions are made into $B$ copies, expanded into $\mathcal{G}_c \in \mathbb{R}^{B \times N \times 3}$, and sampled timestamps are made into $N$ copies, expanded into $\mathcal{T'} \in \mathbb{R}^{B \times N \times 1}$. Then, we composite $\mathcal{G}_c$ and $\mathcal{T'}$ together and apply position encoding function $\gamma$ to extract high frequency information, as in Eq.~\ref{eq: pos_enc}:

\newcommand{\dummypos}{p}
\newcommand{\numfrequencies}{L}

\begin{equation}
\begin{aligned}
    \gamma(\dummypos) = & \left( \sin\left(2^0 \pi \dummypos\right), \cos\left(2^0 \pi \dummypos\right), \right. \cdots \\
    & \left.  \sin\left(2^{\numfrequencies-1} \pi \dummypos\right), \cos\left(2^{\numfrequencies-1} \pi \dummypos\right) \right)
\end{aligned}
\label{eq: pos_enc}
\end{equation}
In our experiments, we set $L=6$ for both positions $x$ and time $t$. We treat $[\gamma(\mathcal{G}_c, \gamma(\mathcal{T'})] \in \mathbb{R}^{B \times N \times (3\times 2\numfrequencies+ 2\numfrequencies) }$ as the original input $F_0$ to \name. It can be considered as $N$ sequences of the length $B$, containing $8\numfrequencies$ feature channels.
With $M$ transformer encoder layers, for $m^{th}$ layer, intermediate features are calculated through multi-head self-attention(MSA) and MLP blocks.
In the final stage, we use a tiny MLP to transform the last encoded features $F_{M-1}$ into offset $\mathcal{O} \in \mathbb{R}^{B \times N \times 3}$ in the linear space:
\begin{equation}
\begin{aligned}
\mathcal{O}=MLP(F_{M-1}), \quad \mathcal{G}_t = \mathcal{G}_c + \mathcal{O}
\end{aligned}
\label{eq: offset}
\end{equation}
We consider the output from the \encoder as a fixing residual term to the original positions to encourage a gradual, steady learning process on motion patterns. 
Implicit cross-time relationship learning in \name enables the automatic aggregation of Gaussians with similar variations during optimization, promoting more efficient spatial distribution and accelerating rendering speed.

\noindent \textbf{Gradient Analysis}
Let $\mathcal{D}$ and $\mathcal{P}$ be the deformation field and \name respectively, the output of the deformation field $\Delta \mu_i$ and partial derivative for \( \mu \) in time $t_i \in \mathcal{T}_s$ are formulated as follows:
\begin{equation}
    \Delta \mu_i = \mathcal{D}(\mu, t_i), \quad \frac{\partial \Delta \mu_i}{\partial \mu} = \frac{\partial \mathcal{D}(\mu, t_i)}{\partial \mu}
\label{eq: backward_ori}
\end{equation}

With \name applied on time batch $\mathcal{T}_s$, we reformulate as Eq.~\ref{eq: forward_ours} and data flow changes as in Fig.~\ref{fig: flow}.
\begin{equation}
    \Delta \mu_i = \mathcal{D}(\mu + \mathcal{P}(\mu, \mathcal{T}_s), t_i)
    \label{eq: forward_ours}
\end{equation}

To calculate the partial derivative for $\mu$, we apply the chain rule. Let $a = \mu + \mathcal{P}(\mu, \mathcal{T}_s)$, then:
\begin{equation}
    \frac{\partial \Delta \mu_i}{\partial \mu} = \frac{\partial \mathcal{D}(a, t_i)}{\partial a} \cdot \frac{\partial a}{\partial \mu}, \quad \frac{\partial a}{\partial \mu} = 1 + \frac{\partial \mathcal{P}(\mu, \mathcal{T}_s)}{\partial \mu}
\end{equation}

\begin{equation}
    \frac{\partial \Delta \mu_i}{\partial \mu} = \frac{\partial \mathcal{D}(a, t_i)}{\partial a} \cdot \left(1 + \frac{\partial \mathcal{P}(\mu, \mathcal{T}_s)}{\partial \mu}\right)
    \label{eq: backward_ours}
\end{equation}

Eq.~\ref{eq: backward_ori} has a weaker dynamic nature, as it only considers the current timestamp $t_i$.
In contrast, Eq.~\ref{eq: backward_ours} incorporates an additional gradient term, $\frac{\partial \mathcal{P}(\mu, \mathcal{T}_s)}{\partial \mu}$, enabling the current state to be influenced by any past or future states. In other words, $\Delta \mu_i$ also optimizes the models according to other timestamps $t_j \in \mathcal{T}_s (j \neq i)$, which is a significant difference compared to the backward process without \name. Such design accounts for cross-time attention and allows the models to capture more challenging motion patterns from a global view of the entire time series.


\subsection{Shared Deformation Fields}
\label{sec: shared}
Accounting for additional computation costs of \name, which can significantly decrease rendering speed, we force the original deformation field and the auxiliary deformation field to share weights for knowledge transferring. 
Let $\mathcal{V}$ be the camera viewpoints, we apply the shared deformation fields to predict deformed space from both Gaussians in the canonical space and Gaussians from \name.
Then, we apply the splatting algorithm to these two groups of deformed space. We calculate the losses between rendered images and ground truth $\mathcal{I}_{gt}$ as follows:
\begin{equation}
 \mathcal{L}_c = \Vert Splatting(\mathcal{D}(\mathcal{G}_c, \mathcal{T}) , \mathcal{V}) -\mathcal{I}_{gt} \Vert_1
 \end{equation}
 \begin{equation}
 \mathcal{L}_t = \Vert Splatting(\mathcal{D}(\mathcal{G}_t, \mathcal{T}) , \mathcal{V}) -\mathcal{I}_{gt} \Vert_1
  \end{equation}
 \begin{equation}
 \mathcal{L} = \lambda_c \mathcal{L}_c + \lambda_t \mathcal{L}_t
\end{equation}
, where $\mathcal{L}_c, \mathcal{L}_t$ represent losses of original branch and \name branch with $\lambda_c > \lambda_t$. We use a relatively smaller $\lambda_t$ because we find it easy to overfit on the second branch with \name, causing a degradation in inference quality.



%% file: sec/5_experiment.tex
\begin{table*}[tbp]
\centering
\tabcolsep=0.1cm

\begin{tabular}{c|cccccccccccc|cc}
\hline
\multirow{2}{*}{Method} & \multicolumn{2}{c}{Sear Steak}  & \multicolumn{2}{c}{Flame Salmon} & \multicolumn{2}{c}{Cut Beef}    & \multicolumn{2}{c}{Flame Steak} & \multicolumn{2}{c}{Cook Spinach} & \multicolumn{2}{c|}{Coffee Martini} & \multicolumn{2}{c}{Mean}         \\
                        & PSNR           & SSIM           & PSNR            & SSIM           & PSNR           & SSIM           & PSNR           & SSIM           & PSNR            & SSIM           & PSNR             & SSIM            & PSNR           & SSIM           \\ \hline
K-Plane~\cite{kplanes_2023}                    & 32.52          & 0.971          & 30.44           & 0.942          & 31.82           & 0.965          & 32.38          & 0.970                   & 30.60            & 0.968   & 29.99           & 0.943         & 31.63           & 0.960          \\
\hline
MixVovels~\cite{wang2022mixed}                    & 31.21          & 0.971          & 29.92           &  0.945         & 31.30           &  0.965         & 31.43          &      0.970              & 31.61            & 0.965   & 29.36           &    0.946     & 30.80           & 0.960          \\
\hline
GS4D~\cite{yang2023gs4d}                    & 32.92          & 0.953          & 26.39           & 0.897          & 33.08           & 0.959          & 33.81          & 0.967                   & 32.77            & 0.956   & 25.23           & 0.884         & 30.07           & 0.936          \\
\hline
4D-Rotor~\cite{4drotorgs}                    & 32.86          & 0.956          & 28.25           & 0.913          & 33.14           & 0.952          & 31.61          & 0.953                   & 32.56            & 0.949   & 27.95           & 0.908         & 31.06           & 0.938          \\
\hline
4DGS~\cite{Wu_2024_CVPR}                                         & 32.49          & 0.949          & 28.92           & 0.917          & 32.90           & 0.956          & 32.51          & 0.954          & 32.46          & 0.948                               & 27.34            & 0.903           & 31.10           & 0.938          \\
+\name                                        & \textbf{33.38} & \textbf{0.955} & \textbf{29.33}  & \textbf{0.924} & \textbf{33.11} & \textbf{0.957} & \textbf{33.25} & \textbf{0.953} & \textbf{33.03} & \textbf{0.949}                      & \textbf{28.93}   & \textbf{0.910}   & \textbf{31.84} & \textbf{0.941}
          \\ \hline
STGS~\cite{Li_STG_2024_CVPR}             & 33.71          & 0.962          & 28.21           & 0.921          & 33.52          & 0.958          & 33.46          & 0.963          & 33.13           & 0.955          & 27.71            & 0.915           & 31.62          & 0.946          \\
+\name          & \textbf{34.34} & \textbf{0.965} & \textbf{29.13}  & \textbf{0.924} & \textbf{33.57} & \textbf{0.958} & \textbf{34.04} & \textbf{0.964} & \textbf{33.45}  & \textbf{0.956} & \textbf{28.83}   & \textbf{0.917}  & \textbf{32.23} & \textbf{0.947} \\ \hline
\end{tabular}
\vspace{-2mm}
\caption{Quantitative Comparisons on N3DV Dataset~\cite{li2022nv3d}. We use \textbf{bold font} to indicate the improvement, statistics of K-Plane~\cite{kplanes_2023} and MixVoxels~\cite{wang2022mixed} are from their original paper, while we calculate metrics of all Gaussian-based methods by running their official codes.}
\label{tab: nv3d_table}
\end{table*}

\begin{figure*}[tbp]
    \centering
    \includegraphics[width=1\linewidth, trim={0 10 0 0}, clip]{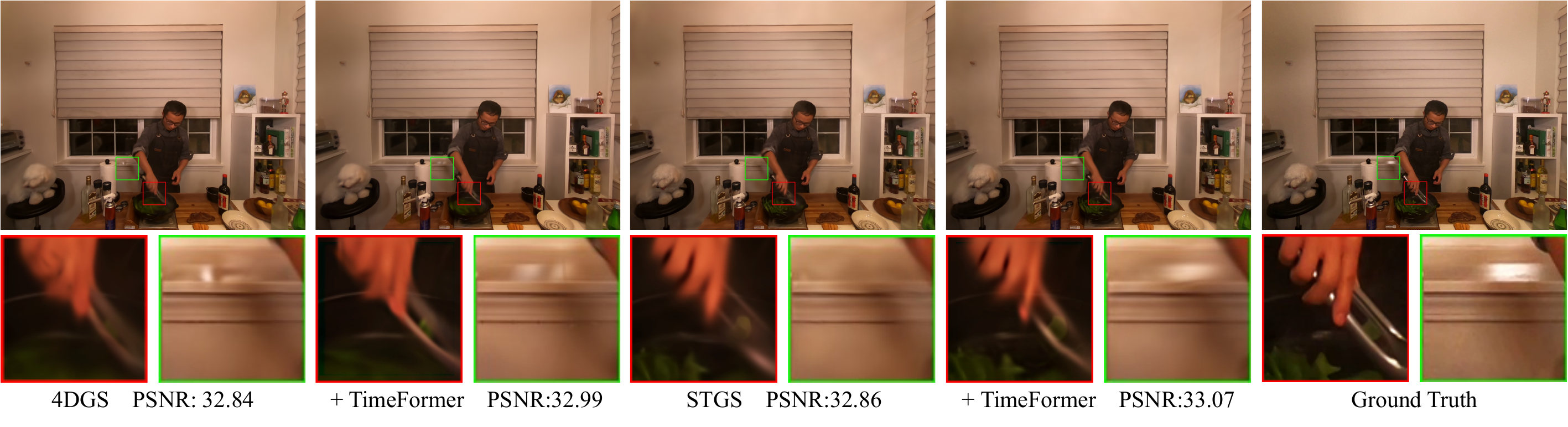}
    \caption{Visualization of Comparisons on N3DV Dataset~\cite{li2022nv3d}. }
    \label{fig: dynerf_compare}
\end{figure*}

\begin{figure}[tbp]
    \centering
    \includegraphics[width=0.9\linewidth]{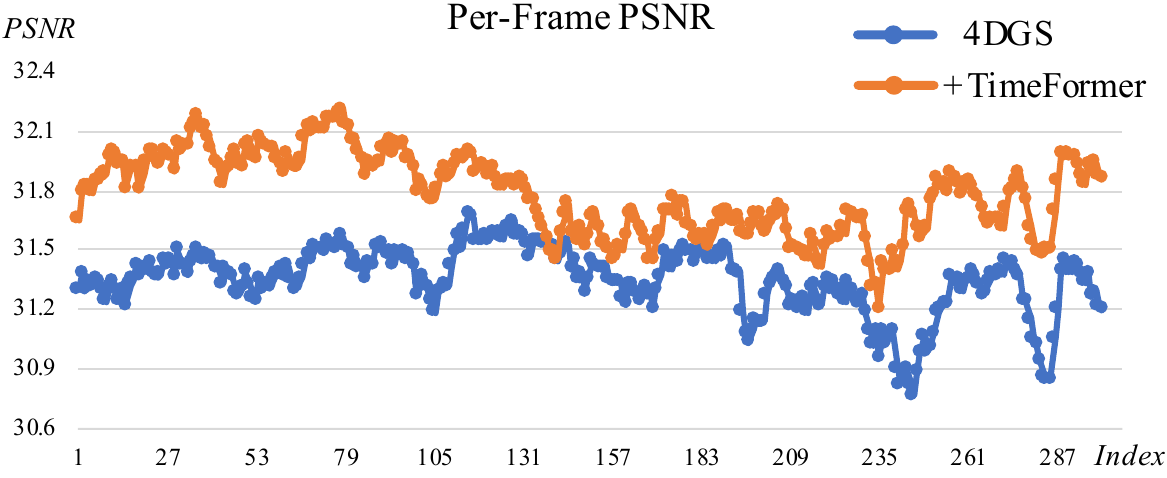}
    \caption{Per-Frame PSNR on N3DV Dataset~\cite{li2022nv3d}. \name improves the reconstruction quality, especially at the beginning and end time series.}
    \label{fig: perframe}
\end{figure}

\begin{figure}[tbp]
    \centering
    \includegraphics[width=0.9\linewidth]{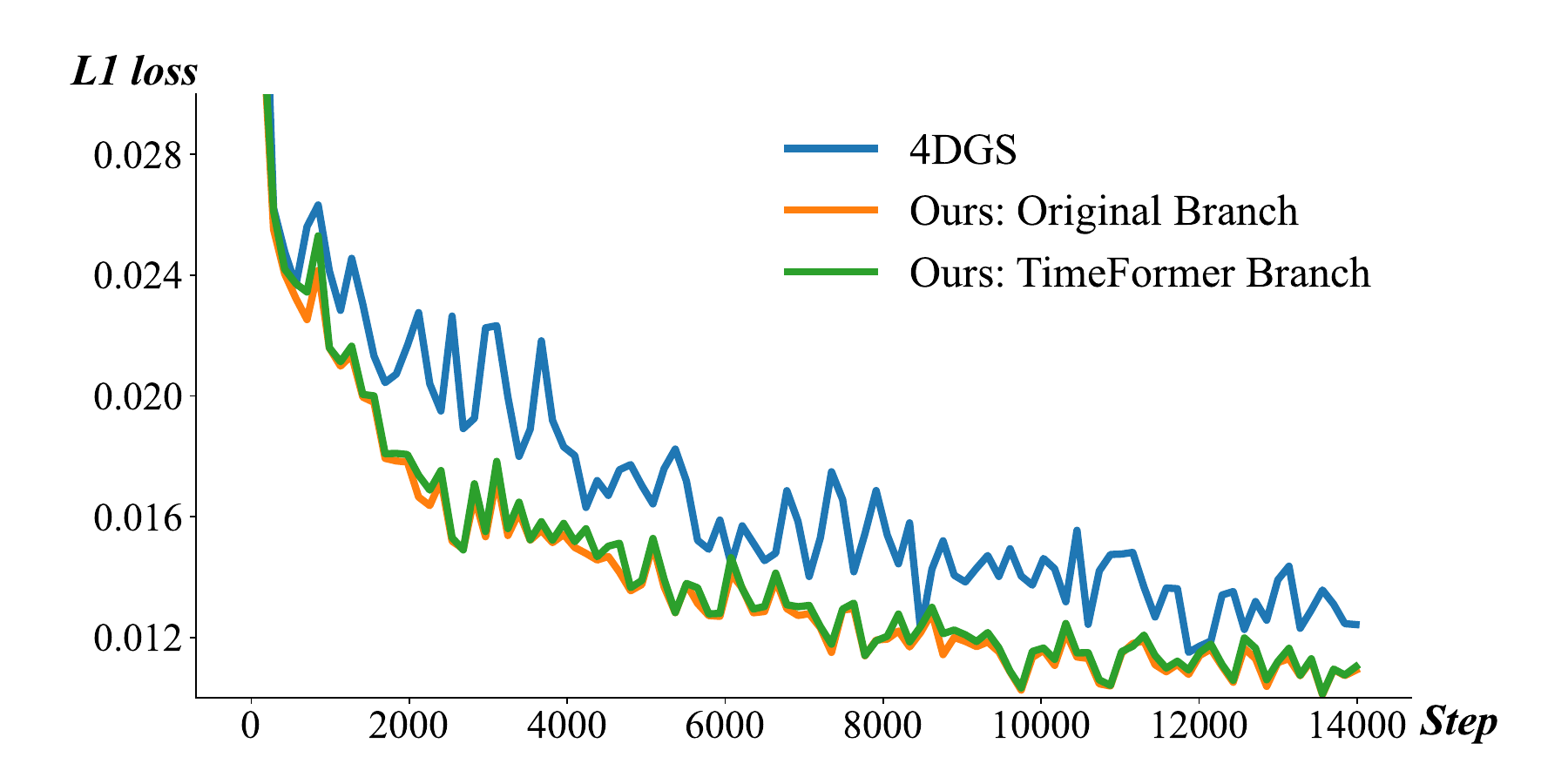}
    \caption{Comparions of Convergence Speed. We use the same batch size for 4DGS~\cite{Wu_2024_CVPR} and calculate the loss of both branches in our method. 
    }
    \label{fig: convergence}
\end{figure}

\begin{table}[tbp]
    \centering
    \begin{subtable}[t]{0.45\textwidth}
        \centering
        \begin{tabular}{p{2.2cm}|ccc}
        \hline
        \centering Method & $N$ $\downarrow$  & Training $\downarrow$ &FPS $\uparrow$  \\ \hline
        \centering STGS~\cite{Li_STG_2024_CVPR}   & 172.1 k &    \textbf{41 min} & 82.1 \\
         \centering+\name  & \textbf{148.9 k}   & 78 min    & \textbf{88.9} \\ \hline
         \centering4DGS~\cite{Wu_2024_CVPR}   & 145.9 k  &  \textbf{52 min}   & 31.7 \\
         \centering+\name  & \textbf{113.1 k}     & 94 min  & \textbf{37.9} \\ \hline
        \end{tabular}
        \caption{Quantitative comparisons on N3DV Dataset~\cite{li2022nv3d}.}
        \vspace{2mm}
    \end{subtable}
    \hfill
    \begin{subtable}[t]{0.45\textwidth}
        \centering
        \begin{tabular}{p{2.2cm}|ccc}
        \hline
        \centering Method & $N$ $\downarrow$  & Training $\downarrow$ &FPS $\uparrow$  \\ \hline
        \centering DeformGS~\cite{yang2023deformable3dgs}   & 169.6 k   & \textbf{25 min} & 30.1 \\
        \centering +\name  & \textbf{82.9 k}    & 35 min    & \textbf{58.9} \\ \hline
        \centering 4DGS~\cite{Wu_2024_CVPR}   & 172.3 k    & \textbf{32 min}   & 35.7 \\
        \centering +\name  & \textbf{135.5 k}     & 48 min  & \textbf{40.9} \\ \hline
        \end{tabular}
        \caption{Quantitative comparisons on HyperNeRF Dataset~\cite{park2021hypernerf}.}
    \end{subtable}
    \caption{Comparisons of Gaussian Number, Training Time \& FPS.}
    \label{tab: point_fps}
\end{table}

\begin{table*}[tbp]
\centering
\tabcolsep=0.11cm

\begin{tabular}{c|cccccccccccc|cc}
\hline
\multirow{2}{*}{Method}                                      & \multicolumn{2}{c}{Broom}       & \multicolumn{2}{c}{Chicken}     & \multicolumn{2}{c}{Cut Lemon}   & \multicolumn{2}{c}{Torchco}     & \multicolumn{2}{c}{Peel Banana} & \multicolumn{2}{c|}{Hand}       & \multicolumn{2}{c}{Mean}         \\
                                                             & PSNR           & SSIM           & PSNR           & SSIM           & PSNR           & SSIM           & PSNR           & SSIM           & PSNR           & SSIM           & PSNR           & SSIM           & PSNR           & SSIM           \\ \hline
DeformGS~\cite{yang2023deformable3dgs} & 20.74          & 0.322          & 26.32          & 0.786          & 27.94          & 0.714          & 27.4           & 0.877          & 26.38          & 0.836          & 27.79          & 0.78           & 26.09          & 0.719          \\ 
+\name                                                        & \textbf{21.01} & \textbf{0.324} & \textbf{26.55} & \textbf{0.792} & \textbf{30.01} & \textbf{0.780} & \textbf{27.55} & \textbf{0.883} & \textbf{27.24} & \textbf{0.852} & \textbf{29.79} & \textbf{0.848} & \textbf{27.03} & \textbf{0.747} \\ \hline
4DGS\cite{Wu_2024_CVPR}             & 21.53          & 0.351          & 26.82          & 0.797          & 29.72          & 0.763          & 27.45          & 0.883          & 27.82          & 0.844          & 29.52          & 0.841          & 27.14          & 0.747          \\
+\name                                                        & \textbf{22.84} & \textbf{0.401} & \textbf{27.06} & \textbf{0.801} & \textbf{30.23} & \textbf{0.778} & \textbf{29.48} & \textbf{0.902} & \textbf{28.21} & \textbf{0.853} & \textbf{30.38} & \textbf{0.862} & \textbf{28.03} & \textbf{0.766} \\ \hline
\end{tabular}
\caption{Quantitative Comparisons on HyperNeRF Dataset~\cite{park2021hypernerf}. 
}
\label{tab: hypernerf_table}

\end{table*}

\begin{figure*}[tbp]
    \centering
    \includegraphics[width=1\linewidth]{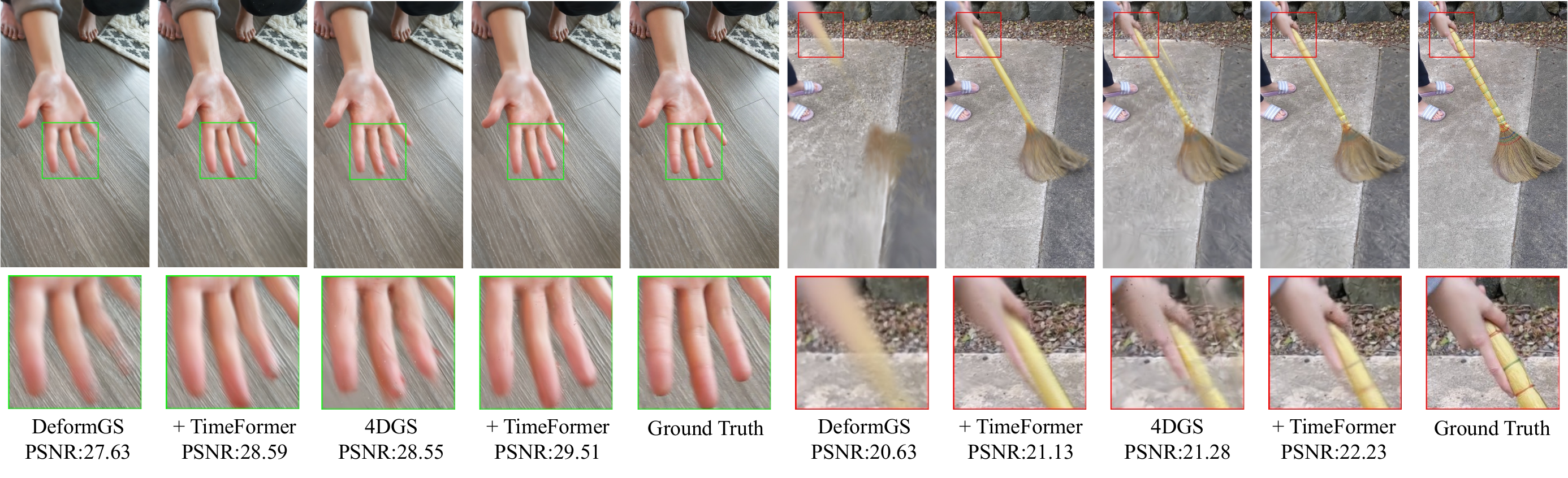}
    \vspace{-8mm}
    \caption{Visualization of Comparisons on HyperNeRF Dataset~\cite{park2021hypernerf}.}
    \label{fig: hypernerf_compare}
\end{figure*}

\begin{table*}[tbp]

\centering
\tabcolsep=0.11cm
\begin{tabular}{c|cccccccccccc|cc}
\hline
\multirow{2}{*}{Method} & \multicolumn{2}{c}{Press}      & \multicolumn{2}{c}{Plate}       & \multicolumn{2}{c}{Basin}       & \multicolumn{2}{c}{Sieve}       & \multicolumn{2}{c}{Bell}        & \multicolumn{2}{c|}{Cup}        & \multicolumn{2}{c}{Mean}       \\
                        & PSNR          & SSIM           & PSNR           & SSIM           & PSNR           & SSIM           & PSNR           & SSIM           & PSNR           & SSIM           & PSNR           & SSIM           & PSNR        & SSIM            \\ \hline
HyperNeRF~\cite{park2021hypernerf}               & 25.4          & 0.873          & 18.1           & 0.714          & 20.2           & 0.829          & 25.0             & 0.909          & 24.0             & 0.884          & 24.1           & 0.896          & 22.8        & 0.851          \\ \hline
NeRF-DS~\cite{yan2023nerf}& 26.4          & 0.911          & 20.8           & 0.867          & 20.3           & 0.868          & 26.1           & 0.935          & 23.3           & 0.872          & 24.5           & 0.916          & 23.57       & 0.895          \\ \hline
DeformGS\cite{yang2023deformable3dgs} & 25.68          & 0.866          & 20.82          & 0.812          & 19.87          & 0.804          & 25.71          & 0.881          & 24.92          & 0.854          & 24.52          & 0.897          & 23.59          & 0.852          \\
+\name                                                        & \textbf{26.29} & \textbf{0.867} & \textbf{20.90} & \textbf{0.817} & \textbf{19.93} & \textbf{0.805} & \textbf{26.26} & \textbf{0.891} & \textbf{25.90} & \textbf{0.873} & \textbf{25.16} & \textbf{0.903} & \textbf{24.10} & \textbf{0.859}
 \\ \hline
\end{tabular}
\caption{Quantitative Comparisons on NeRF-DS Dataset~\cite{yan2023nerf}. 
}
\label{tab: nerfds_table}
\end{table*}

\begin{figure*}[tbp]
    \centering
    \includegraphics[width=\linewidth]{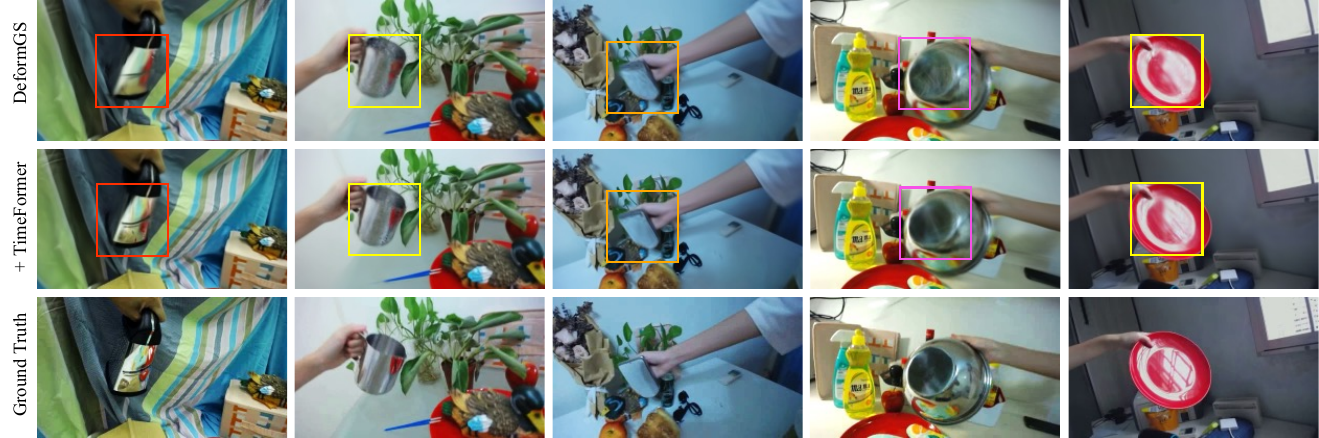}
    \caption{Visualization of Comparisons on NeRF-DS Dataset~\cite{yan2023nerf}. Compared with original results from DeformGS~\cite{yang2023deformable3dgs}, \name presents a clearer visual effect on dynamic objects with specular surfaces.}
    \label{fig: nerfds_compare}
\end{figure*}

\begin{figure}[tbp]
    \centering
    \includegraphics[width=1\linewidth]{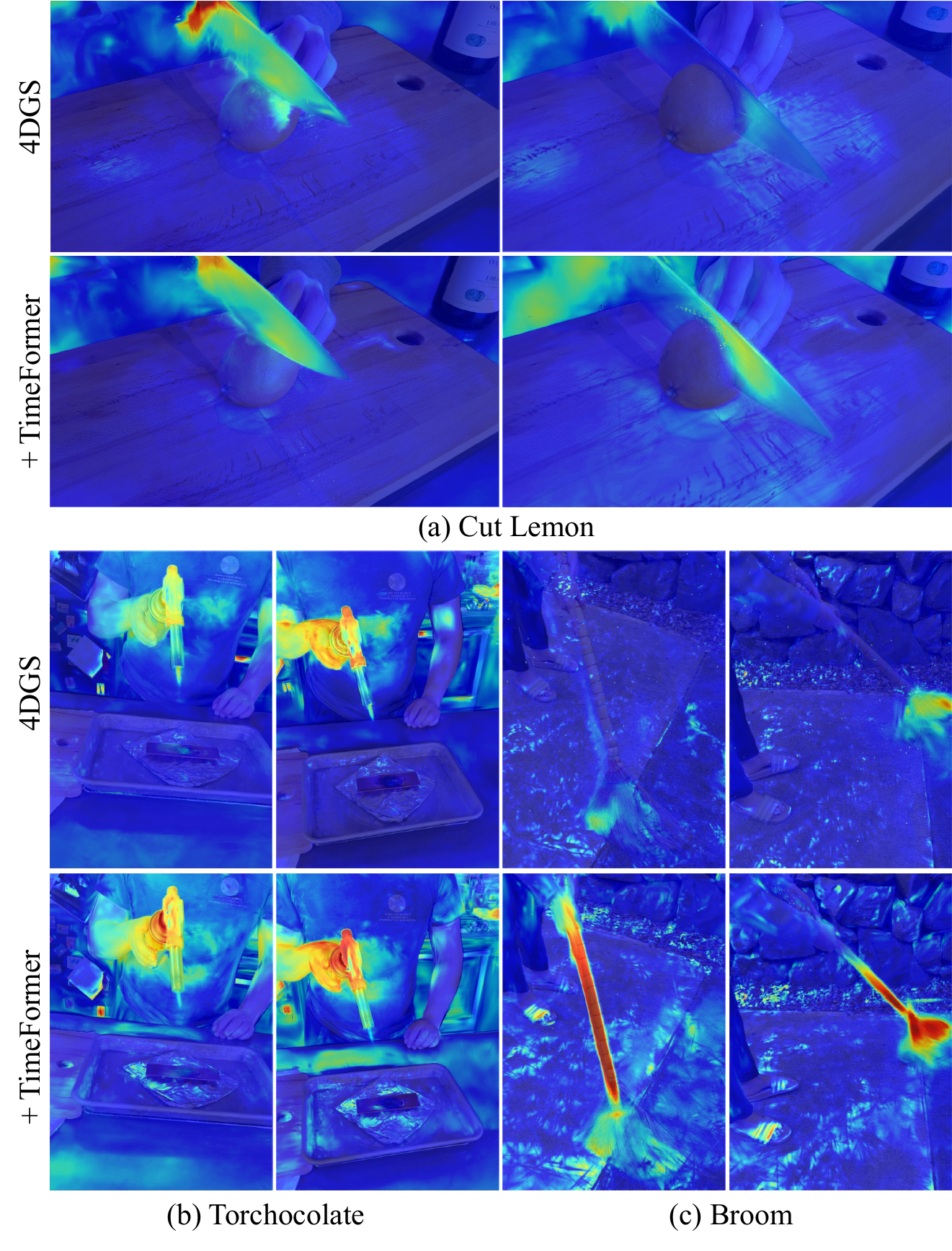}
    \caption{Motion Visualization on HyperNeRF Dataset~\cite{park2021hypernerf}.
    }
    \label{fig: motion_vis}
    \vspace{-5mm}
\end{figure}

\section{Experiment}

\subsection{Implementation Details}
\label{sec: impl}
Our implementation is tested on a single A100 GPU. We use $M=4$ transformer encoder layers in \name and the number of time samples is $B=4$. We set $\lambda_c = 1$ and $\lambda_t = 0.8$ to prevent overfitting on \name. We assess our experimental results using image quality metrics, including peak-signal-to-noise ratio (PSNR) and structural similarity index (SSIM~\cite{ssim}). 

\noindent \textbf{Baselines \& Datasets. } We apply \name to 4DGS~\cite{yang2023gs4d}, STGS~\cite{Li_STG_2024_CVPR} on multi-view videos, \eg, N3DV Dataset~\cite{li2022nv3d}, and evaluate \name on 4DGS~\cite{yang2023gs4d} and DeformGS~\cite{yang2023deformable3dgs} on monocular dynamic videos, \eg, HyperNeRF Dataset~\cite{park2021hypernerf}, respectively. We also utilize NeRF-DS Dataset~\cite{yan2023nerf} to demonstrate the robustness of \name on moving objects with specular surfaces.

\subsection{Experimental Comparisons}
\label{sec: results}

\noindent \textbf{Multi-View Video Dataset.} 
In Tab.~\ref{tab: nv3d_table}, \name improves the reconstruction quality (\eg, PSNR) of original 4DGS~\cite{Wu_2024_CVPR} and STGS~\cite{Li_STG_2024_CVPR} by 0.74 and 0.61, outperforming all baselines methods. We extract the frames at an initial time for scene \textit{Cook Spanish} in Fig.~\ref{fig: dynerf_compare}, and \name can reconstruct a clearer geometry and specular effects compared to the original results.

To have a more comprehensive overview of \name's capacity on different temporal stages, we collect PSNR for 300 frames of all six scenes from N3DV Dataset~\cite{li2022nv3d} and calculate the average per-frame PSNR. As in Fig.~\ref{fig: perframe}, the reconstruction quality increases more significantly at the beginning and end, and such balanced reconstruction results are attributed to \name's cross-temporal attention mechanism on the whole time series.



\noindent \textbf{Monocular Video Dataset.} Tab.~\ref{tab: hypernerf_table} shows the enhancements achieved by \name on the HyperNeRF Dataset. It's observed that \name increases the PSNR by 0.89 for 4DGS~\cite{Wu_2024_CVPR} and 0.94 for DeformGS~\cite{yang2023deformable3dgs}, and the SSIM by 0.019 and 0.028, respectively. Fig.~\ref{fig: hypernerf_compare} additionally provides visualization of the improvement in the image quality: on the left, \name generates a clearer outline between fingers, while on the right the texture on the front of the broom is clearer. \name also eliminates artifacts and reveals a clear structure of the broom.

\noindent \textbf{Dynamic Reflective Video Dataset.} 
Fig.~\ref{fig: nerfds_compare} additionally demonstrates that \name can work well on scenes with dynamic specular objects on NeRF-DS Dataset~\cite{yan2023nerf}. 
\name can reconstruct a cleaner appearance on reflective material (\eg, a metallic cup, a glass bottle), and this improvement is attributed to the attention mechanism which enables the deformation field to detect changes on specular surfaces from a learning perspective automatically.

\noindent \textbf{Analysis of FPS \& Canonical Space. }
We also find that \name achieves more efficient representation with much fewer points in the final canonical space, thereby achieving a higher FPS in Tab.~\ref{tab: point_fps}. 
We attribute the phenomenon to the reason that \name guides faster gradient descent and promotes the spatial layout of Gaussian points during optimization. Interestingly, in Fig.~\ref{fig: convergence}, the L1 loss of both branches in our method decreases faster and shows almost the same convergence trend during training, exceeding the optimization speed of the original method.
This eliminates many redundant points and guides the canonical space toward a more efficient distribution, as in Fig.~\ref{fig: point_cloud}. 
Moreover, Fig.~\ref{fig: convergence} also proves that the cross-time relationship learned in \name has been successfully transferred to the base branch, as introduced in Sec.~\ref{sec: shared}.

\noindent \textbf{Analysis of Motion Patterns. } We argue that \name achieves more robust learning of motion patterns. To take a deeper insight, we visualize the motion changes as follows: for each Gaussian, we calculate its bias $(\Delta_x, \Delta_y, \Delta_z)$ towards canonical space in time $t$ and replace the color attributes $(r,g,b)$ with the absolute value of this bias: 
$
    (r, g, b) \leftarrow (\lvert \Delta_x \rvert, \lvert \Delta_y \rvert, \lvert \Delta_z \rvert) 
$.
We use the same splatting algorithm to render the accumulated motion bias, transferring the rendered result as a heatmap, as in Fig.~\ref{fig: motion_vis}.

In Fig.~\ref{fig: motion_vis}\textcolor{cvprblue}{a}, \name identifies the blade’s motion as a rigid deformation, showing a sharper outline and consistent motion, especially as it contacts the initially static lemon. This demonstrates \name’s ability to distinguish between moving and static objects. In Fig.~\ref{fig: motion_vis}\textcolor{cvprblue}{b}, \name captures subtle motion on the spray gun and clear flame flickering on the beef. Fig.~\ref{fig: motion_vis}\textcolor{cvprblue}{c} shows that \name also detects the violent movement of the elongated broom, where 4DGS~\cite{Wu_2024_CVPR} fails, proving its robustness in handling extremely geometries.

\subsection{Ablation Studies}

\noindent \textbf{\encoder. } Tab.~\ref{tab: ablation_tab} demonstrates that the performance of \name is not sensitive to time batch $B$ and transformer encoder layer $M$. In Tab.~\ref{tab: abla_layer_share}, we explore two sampling methods: random sampling and continuous sampling of timestamps. Both strategies demonstrate improvements in PSNR and other metrics compared to the baseline. Notably, the random sampling method yields more significant enhancements. We hypothesize that this phenomenon occurs because the random sampling method allows for more effective propagation of dynamic behavior modeling across the entire time series.

\noindent \textbf{Shared Deformation Fields. } In Tab.~\ref{tab: abla_layer_share}, the image quality during inference suffers a considerable decrease if we train two deformation fields without sharing their weights. Fig.~\ref{fig: convergence} also demonstrates the effectiveness of the two-stream strategy and \name's ability of re-balancing.

\begin{table}[tbp]
    \centering
    \begin{subtable}[t]{0.5\textwidth}
        \centering
        \tabcolsep=0.08cm
        \begin{tabular}{c|ccc|ccc}
\hline
\multirow{2}{*}{Setting} & \multicolumn{3}{c|}{Random Sampling}                                             & \multicolumn{3}{c}{Continuous Sampling}                                         \\
                         & \multicolumn{1}{l}{PSNR$\uparrow$} & \multicolumn{1}{l}{SSIM$\uparrow$} & \multicolumn{1}{l|}{LPIPS$\downarrow$} & \multicolumn{1}{l}{PSNR$\uparrow$} & \multicolumn{1}{l}{SSIM$\uparrow$} & \multicolumn{1}{l}{LPIPS$\downarrow$} \\ \hline
$B$=2                  & 25.89                    & 0.875                    & 0.117                      & 25.73                    & 0.871                    & 0.124                     \\
$B$=4                  & 26.06                    & 0.871                    & 0.119                      & 25.78                    & 0.874                    & 0.117                     \\
$B$=6                  & 25.99                    & 0.87                     & 0.123                      & 25.73                    & 0.865                    & 0.125                     \\ \hline
\end{tabular}
        \caption{Ablation Studies on Batch Size and Sampling Strategies. 
        }
        \label{tab: abla_batch_sample}
        \vspace{2mm}
    \end{subtable}
    \hfill
    \begin{subtable}[t]{0.5\textwidth}
        \centering
        \begin{tabular}{c|ccc}
\hline
Setting    & \multicolumn{1}{l}{PSNR$\uparrow$} & \multicolumn{1}{l}{SSIM$\uparrow$} & \multicolumn{1}{l}{LPIPS$\downarrow$} \\ \hline
Baseline                 & 25.44  & 0.867  & 0.125  \\ \hline
M=1                      & 25.96  & 0.874  & 0.117  \\
M=2                      & 25.90  & 0.874  & 0.120  \\
M=3                      & 26.05  & 0.876  & 0.112  \\ 
M=4                      & 26.04  & 0.873  & 0.119  \\ \hline
w/o Shared & 24.81                    & 0.852                    & 0.132                     \\ \hline
\end{tabular}
        \caption{Ablation Results. $M$ is the number of encoder layers, ``w/o Shared'' means not using shared weights in two deformation fields.}
        \label{tab: abla_layer_share}
    \end{subtable}
    \caption{Ablation Results. The results are from three scenes \textit{press}, \textit{sieve} and \textit{bell} on NeRF-DS Dataset~\cite{yan2023nerf}.}
    \label{tab: ablation_tab}
    \vspace{-5mm}
\end{table}

%% file: sec/6_conclusion.tex
\section{Conclusion}
We propose \name, a Transformer module that is plug-and-play to existing deformable 3D Gaussians methods and enhances reconstruction results without additional computational budget. \name enables deformation fields to implicitly model complex motion patterns from a learning perspective. In addition, we design a two-stream optimization strategy to transfer the learned motion knowledge from \name to the original deformation branch. This allows us to remove \name during inference and thus maintain the same inference speed as the original methods. Extensive experiments demonstrate \name effectively facilitates the reconstruction of three state-of-the-art deformable 3D Gaussians Splatting methods among three datasets, and illustrate the improvement on reconstructing complex scenes containing violent movement, extreme-shaped geometries, or reflective surfaces.

\noindent\textbf{Limitations} \name may produce overly smooth textures for objects with intricate details.

%% file: sec/X_suppl.tex

\appendix

\section{Implementation}
\label{sec:appendix_implementation}
In this section, we provide the Pytorch~\cite{paszke2019pytorch} code of \name in Alg.~\ref{alg: timeformer_code} and two-stream optimization strategy in Alg.~\ref{alg: two_stream}, to clarify the framework in Fig.~\ref{fig: framework}.

Alg.~\ref{alg: timeformer_code} is an additional explanation for Fig.~\ref{fig: gs_timeformer}, including three parts: 1) Position Encoding (PE), 2) Definition of Transformer encoder, 3) Definition of tiny MLP. Note that we use a shared \name on all Gaussians in the canonical space.

The Transformer Encoder receives input structured as [seq\_len, seq\_batch, channel], and we input x structured as [$B$, $N$, 4], where $B$ is the size of time batch, $N$ is the number of Gaussians and 4 means 3 position channel and 1 channel for the timestamp. Alg.~\ref{alg: two_stream} shows how we construct input to time.

\begin{algorithm}[tbp]
  \algsetup{linenosize=\tiny}
  \scriptsize
\SetAlFnt{\tiny}
\SetAlCapFnt{\small}
\SetAlCapNameFnt{\small}
\SetAlgoLined
\PyCode{class \name(nn.module):} \\
\Indp
\PyComment{d\_in: here is 4, (x, y, z, t)} \\
\PyComment{L: frequency of Position Encoding (PE) $\gamma$} \\
\PyCode{def \_\_init\_\_(self, d\_in, L, nhead, d\_hidden, n\_layer):} \\
\Indp
\PyComment{PE: PE function, PE\_ch: d\_in$\times$2L} \\
\PyCode{self.PE, PE\_ch = get\_PE(L=L, d\_in=d\_in)} \\
\PyComment{define \encoder} \\
\PyCode{layer = nn.TransformerEncoderLayer(PE\_ch, nhead, d\_hidden, activation=nn.functional.tanh)} \\
\PyCode{self.encoder = nn.TransformerEncoder(layer, n\_layer)} \\
\PyComment{define Tiny MLP}\\
\PyCode{self.mlp = nn.Linear(PE\_ch, 3)} \\
\Indm

\PyCode{} \\

\PyComment{x: [seq\_len, seq\_batch, channel] }\\
\PyCode{def forward(self, x):} \\
\Indp
\PyCode{PE\_x = self.PE(x)} \\
\PyCode{h = self.encoder(PE\_x)} \\
\PyCode{h = self.mlp(h)} \\
\PyCode{return h} \\
\Indm

\Indm 
\caption{Implementation of \name.}
\label{alg: timeformer_code}
\end{algorithm}

\begin{algorithm}[tbp]
  \algsetup{linenosize=\tiny}
  \scriptsize
\SetAlFnt{\tiny}
\SetAlCapFnt{\small}
\SetAlCapNameFnt{\small}
\SetAlgoLined

\PyComment{timeformer: TimeFormer} \\
\PyComment{deform: Shared Deformation Field} \\
\PyCode{} \\
\PyComment{N: number of Gaussians} \\
\PyComment{GS: Gaussians in the canonical space} \\
\PyComment{B: size of time batch} \\
\PyCode{B = 4}  \\
\PyCode{lambda\_t = 0.8} \\
\PyComment{Vs, Ts: sampled cameras and timestamps} \\
\PyComment{images\_gt: sampled GT images} \\
\PyCode{Vs, Ts, images\_gt = random.sample(Dataset, B)} \\
\PyCode{} \\
\PyComment{construct input to \name} \\
\PyComment{[N, 3] => [B, N, 3]} \\
\PyCode{G\_expanded = GS.xyz.unsqueeze(0).expand(B, -1, -1)} \\
\PyComment{[B] => [B, N, 1]} \\
\PyCode{T\_expanded = Ts.unsqueeze(1).expand(-1, N).unsqueeze(2)}

\PyComment{src: [B(seq\_len), N(seq\_batch), 4(channel)]} \\
\PyCode{src = torch.cat((G\_expanded, T\_expanded), dim=2)} \\
\PyComment{offset\_t: [B, N, 3]} \\
\PyCode{offset\_t = timeformer(src)}  \\

\PyCode{} \\

\PyComment{use iterations to save cuda memory} \\
\PyCode{loss = 0.0} \\
\PyCode{for i in range(B):} \\
\Indp
\PyComment{original branch} \\
\PyComment{simplified: (xyz, t) => d\_xyz} \\
\PyCode{d\_xyz = deform(GS.xyz, Ts[i])} \\
\PyCode{image = splatting(GS, Vs[i], d\_xyz=d\_xyz)} \\
\PyCode{loss += L1(image, images\_gt[i])} \\
\PyCode{} \\
\PyComment{\name branch: use offset\_t[i,:,:]} \\
\PyCode{d\_xyz\_t = deform(GS.xyz+offset\_t[i,:,:], Ts[i])} \\
\PyCode{image\_t = splatting(GS, Vs[i], d\_xyz=d\_xyz\_t)} \\
\PyCode{loss += lambda\_t * L1(image\_t, images\_gt[i])} \\
\Indm
\PyCode{} \\
\PyComment{Optimize shared deformation field, \name and Gaussians in the canonical space} \\
\PyCode{loss.backward()} \\

\PyCode{deform.optimizer.step()} \\
\PyCode{timeformer.optimizer.step()} \\
\PyCode{GS.optimizer.step()} \\

\caption{Two-Stream Optimization Strategy.}
\label{alg: two_stream}
\end{algorithm}

In Alg.~\ref{alg: two_stream}, we introduce the process in a time batch in detail. We first construct input to \name by concatenating Gaussian positions and sampled time stamps, as in Sec.~\ref{sec: gs_timeformer}. Besides the original branch where the deformation function is directly performed on the canonical space, we add another \name branch. \name calculates prior offsets ``offset\_t'' via cross-time relationships before the deformation field. These two branches are optimized at the same time during training, while the \name branch can be removed during inference.

\section{Analysis on Canonical Space \& FPS}
\begin{figure*}[tbp]
    \centering
    \includegraphics[width=\linewidth]{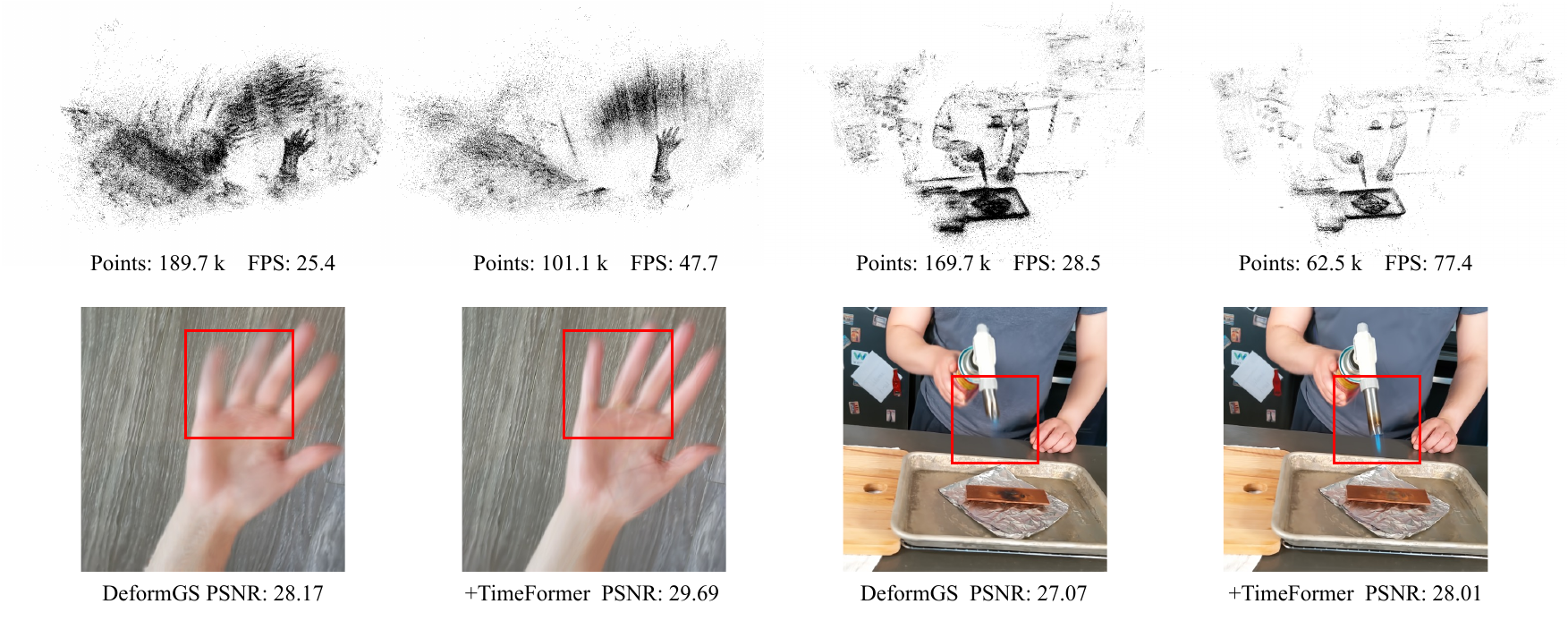}
    \caption{Comparisons of Canonical Space, FPS on Hypernerf Dataset~\cite{park2021hypernerf}.}
    \label{fig: supp_point_hypernerf}
\end{figure*}

\begin{figure*}[!ht]
    \centering
    \includegraphics[width=\linewidth]{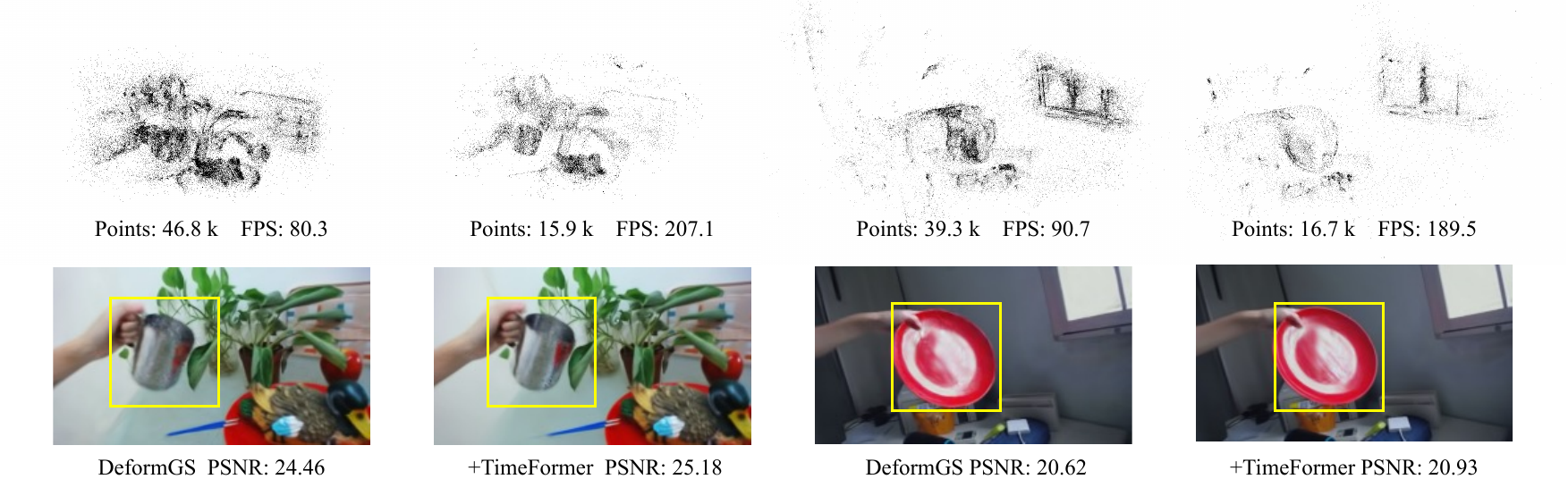}
    \caption{Comparisons of Canonical Space, FPS on NeRF-DS Dataset~\cite{yan2023nerf}.}
    \label{fig: supp_point_nerfds}
\end{figure*}

Apart from Fig.~\ref{fig: point_cloud}, we provide more results in Fig.~\ref{fig: supp_point_hypernerf} and Fig.~\ref{fig: supp_point_nerfds}, as a further illustration on \name's capability to reduce Gaussians in the canonical space and improve inference speed. \name promotes more efficient spatial distribution of Gaussians in the canonical space, leading to improvements in reconstruction quality while simultaneously eliminating a substantial number of redundant Gaussians compared to baseline methods.


\begin{table*}[!ht]
\tabcolsep=0.12cm
\begin{tabular}{c|ccc|ccc|ccc|ccc}
\hline
\multirow{2}{*}{Setting} & \multicolumn{3}{c|}{Bell} & \multicolumn{3}{c|}{Press} & \multicolumn{3}{c|}{Sieve} & \multicolumn{3}{c}{Mean} \\
                         & PSNR$\uparrow$    & SSIM$\uparrow$   & LPIPS$\downarrow$  & PSNR$\uparrow$    & SSIM$\uparrow$    & LPIPS$\downarrow$  & PSNR$\uparrow$    & SSIM$\uparrow$    & LPIPS$\downarrow$  & PSNR$\uparrow$   & SSIM$\uparrow$   & LPIPS$\downarrow$  \\ \hline
Baseline                 & 24.92   & 0.854  & 0.126  & 25.68   & 0.866   & 0.141  & 25.71   & 0.881   & 0.108  & 25.44  & 0.867  & 0.125  \\ \hline
M=1                      & 25.67   & 0.872  & 0.097  & 26.17   & 0.867   & 0.141  & 26.04   & 0.884   & 0.113  & 25.96  & 0.874  & 0.117  \\
M=2                      & 25.46   & 0.872  & 0.104  & 26.01   & 0.866   & 0.139  & 26.22   & 0.883   & 0.116  & 25.90  & 0.874  & 0.120  \\
M=3                      & 25.87   & 0.875  & 0.095  & 26.29   & 0.865   & 0.138  & 26.00   & 0.887   & 0.104  & 26.05  & 0.876  & 0.112  \\
M=4                      & 25.71   & 0.870  & 0.103  & 26.09   & 0.865   & 0.14   & 26.33   & 0.885   & 0.115  & 26.04  & 0.873  & 0.119  \\ \hline
w/o Shared               & 24.12   & 0.832  & 0.137  & 25.01   & 0.854   & 0.146  & 25.29   & 0.869   & 0.113  & 24.81  & 0.852  & 0.132  \\ \hline
\end{tabular}
\caption{Ablation Results of three scenes \textit{press}, \textit{sieve} and \textit{bell} on NeRF-DS Dataset~\cite{yan2023nerf}.}
\label{tab: supp_ablation}

\end{table*}

\section{Slider Window Demo}
We provide additional comparison results of baseline methods with \name. We strongly recommend opening the following website demos in folder \textit{website\_demos} and using the ``slider window'' to see the improvements brought by \name more clearly.

\begin{itemize}
    \item Overview: index.html
    \item 4DGS+\name on HyperNeRF Dataset: index\_4DGS\_hypernerf.html
    \item DeformGS+\name on HyperNeRF Dataset: index\_deformGS\_hypernerf.html
    \item DeformGS+\name on NeRF-DS Dataset: index\_deformGS\_nerfds.html
\end{itemize}

\section{Ablation Studies}
We provide more detailed ablation results on three scenes on NeRF-DS Dataset~\cite{yan2023nerf}, as in Tab.~\ref{tab: supp_ablation}. This further illustrates that \name is not sensitive to the number of transformer encoder layers $M$. However, we observe a significant decrease in reconstruction quality on all scenes without shared deformation fields.